\documentclass[10pt,twocolumn,letterpaper]{article}

\usepackage{wacv}
\usepackage{times}
\usepackage{epsfig}
\usepackage{graphicx}
\usepackage{amsmath}
\usepackage{amssymb}
\usepackage{algorithm}
\usepackage{algpseudocode}

\wacvfinalcopy 

\def\wacvPaperID{517} 

\ifwacvfinal\fi
\begin{document}

\title{Resisting Large Data Variations via Introspective Transformation Network}

\author{Yunhan Zhao$^1$ \quad Ye Tian$^{2,3}$ \quad Charless Fowlkes$^1$ \quad Wei Shen$^2$ \quad Alan Yuille$^2$ \\
$^1$ University of California, Irvine \quad $^2$ Johns Hopkins University\\
$^3$ Tencent Hippocrates Research Labs\\
{\tt\small \{yunhaz5, fowlkes\}@ics.uci.edu, \quad \{ytian27, wshen10, ayuille1\}@jhu.edu}
}

\maketitle
\ifwacvfinal\thispagestyle{empty}\fi

\begin{abstract}
    Training deep networks that generalize to a wide range of variations in test data is essential to building accurate and robust image classifiers. Data variations in this paper include but not limited to unseen affine transformations and warping in the training data. One standard strategy to overcome this problem is to apply data augmentation to synthetically enlarge the training set. However, data augmentation is essentially a brute-force method which generates uniform samples from some pre-defined set of transformations. In this paper, we propose a principled approach named introspective transformation network (ITN) that significantly improves network resistance to large variations between training and testing data. This is achieved by embedding a learnable transformation module into the introspective network, which is a convolutional neural network (CNN) classifier empowered with generative capabilities. Our approach alternates between synthesizing pseudo-negative samples and transformed positive examples based on the current model, and optimizing model predictions on these synthesized samples. Experimental results verify that our approach significantly improves the ability of deep networks to resist large variations between training and testing data and achieves classification accuracy improvements on several benchmark datasets, including MNIST, affNIST, SVHN, CIFAR-10 and miniImageNet.
\end{abstract}

\section{Introduction}
There has been rapidly progress on vision classification problem due to advances in convolutional neural networks (CNNs) \cite{CNN, alexNet, VGG, googleNet, ResNet, huang2017densely}. CNNs are able to produce promising performance given sufficient training data. However, generalization performance is poor when the training data is limited and fails to cover the range of variations in the testing data (e.g., training on MNIST, but testing on distorted digits in affNIST). Consequently, training deep networks that can resist large variations between training and testing data is a significant challenge for producing accurate and robust image classifiers. 

A typical strategy to improve robustness is to apply data augmentation to enlarge the training set by applying various transformations such as random translations, rotations and flips as well as Gaussian noise injection to the available training images. This strategy is often very effective in improving generalization performance, but it is essentially an inefficient, brute-force method that exhaustively samples from all possible transformations for every training sample. Furthermore, the range and type of such augmentations are chosen heuristically with no strong theoretical justification. 

\begin{figure*}[t]
\begin{center}
    \includegraphics[scale=0.40]{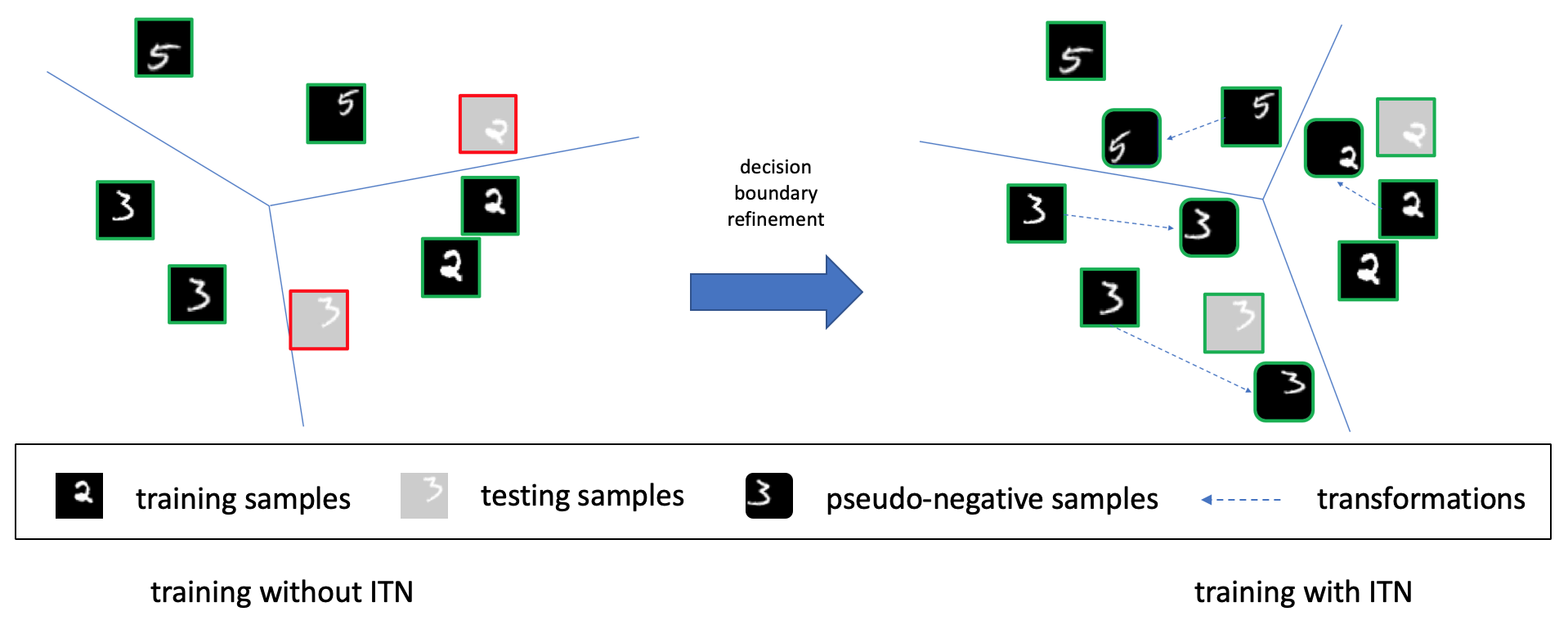}
\end{center}
    \caption{Illustration of the intuition of our ITN framework. ITN enhances the discriminator by generating additional pseudo-negative samples in the training step.}
    \label{fig: ITN }
    \vspace{-10pt}
\end{figure*}

An alternative to augmentation is to synthesize additional training examples using a generative model. {\em How can we automatically generate synthetic samples that are  useful as training data to improve the robustness of CNNs to large variations in testing data?} In this paper, we achieve this by embedding a learnable transformation module into \emph{introspective networks}  (INs)~\cite{ICN,lazarow2017introspective}, a CNN classifier which is also generative. We name our approach \emph{introspective transformation network} (ITN) and we train with a reclassification-by-synthesis algorithm. This alternatively synthesizes samples with learned transformations and enhances the classifier by retraining it with synthesized samples. The intuition of our approach is illustrated in Figure~\ref{fig: ITN }. We use a min-max formulation to learn our ITN, where the transformation module transforms positive samples to maximize their variation from the original training samples and the CNN classifier is updated by minimizing the classification loss of the transformed positive and synthesized pseudo-negative samples. The transformation modules are learned jointly with the CNN classifier, which augments training data in an intelligent manner by narrowing down the search space for the variations.

Our proposed framework is general and can theoretically work with generative adversarial networks (GANs)~\cite{goodfellow2014generative, cgan, wgan, wgan-gp, DCGAN, improved-gan, isola2017image, zhu2017unpaired}. We choose INs in our approach rather than more well known GANs because existing GANs are designed to produce high quality generators while INs have been shown to produce improved discriminators. Additionally, INs can quantitatively prevent generation of bad samples as they have a quality control mechanism in the generation process. Instead of directly decoding the latent vectors in GANs, INs iteratively refine the generated samples until they satisfy the criteria. Moreover, INs avoid adversarial training, which allows them to generate samples more stably than GANs. We verify this choice of generative models (i.e., INs rather than GANs) experimentally. We implement auxiliary classifier GANs (AC-GANs)~\cite{acgan} in our same framework. We adopt AC-GAN since our framework requires the generative models have the ability to generate class conditional samples to enhance the classifier. We term this alternate model {\em auxiliary classifier generative adversarial transformation networks} (AC-GATNs) and qualitatively compare ITN and AC-GATN in Section 4.3, demonstrating the benefit of choosing ITNs for building robust image classifiers.

The main contribution of the paper is a principled approach that endows classifiers with the ability to resist larger variations between training and testing data in an intelligent and efficient manner. ITN enhances the classifier by generating additional training samples, including self-generated pseudo-negative samples and transformed input samples. Experimental results show that our approach achieves better performance than standard data augmentation not only on large variation classification tasks but also maintain strong performance on benchmark datasets. Furthermore, we also show that our approach has great abilities to resist different types of variations between training and testing data.


\section{Related Work}
In recent years, a number of works have emerged that focus on resisting large variations between training and testing data. The most widely adopted approach is data augmentation which applies pre-defined transformations to the training data. Nevertheless, this method is lacking since the user has to specify by-hand the types and extent of transformations applied to the training set. Better methods have been proposed by building connections between generative models and discriminative classifiers~\cite{friedman2001elements, liang2008asymptotic, tu2008brain, jebara2012machine, welling2003self}. These methods capture the underlying generation process of the entire dataset. The discrepancy between training and test data is reduced by generating more samples from the data distribution. 


Development of GANs~\cite{goodfellow2014generative} has led to a huge wave of work in exploring generative-adversarial structures. Combining this structure with deep CNNs can produce models that have stronger generative abilities. In GANs, generators and discriminators are trained simultaneously. Generators try to generate fake images that fool the discriminators, while discriminators try to distinguish the real and fake images. Many variations of GANs have emerged in the past three years, like DCGAN~\cite{DCGAN}, WGAN~\cite{wgan} and WGAN-GP~\cite{wgan-gp}. These GANs variations show stronger learning ability that enables generating complex images. Techniques have been proposed to improve adversarial learning for image generation~\cite{improved-gan, wgan-gp, denton2015deep} as well as for training better image generative models~\cite{DCGAN,isola2017image}. In contrast to work on GANs, here we focus on the complementary problem of using generative models to produce better discriminators.


Introspective networks~\cite{tu2007learning, lazarow2017introspective, ICN, WINN} provide an alternative approach to generate samples. Introspective networks are closely related to GANs since they both have generative and discriminative abilities but differ in several ways. Introspective networks maintain a single model that is both discriminative and generative at the same time while GANs have distinct generators and discriminators. Introspective networks leverage introspective learning which synthesizes samples directly from a discriminative classifier. The generators in GANs provide explicit mappings from a latent space or input to an image while introspective networks represent the distribution of positive images with an efficient sampling/inference process.

Our approach (ITN) utilizes introspective learning, specifically Wasserstein introspective neural networks (WINN). Our contribution is to address the large variations between training and testing data by producing unseen variations using transformers, similar to data augmentation.  However, unlike data augmentation which heuristically samples the space of transformations in an exhaustive way, ITN utilizes a natural min-max problem that searches samples more efficiently and effectively than standard data augmentation.

\section{Method}
We first briefly review the Wasserstein introspective neural networks (WINN) proposed by~\cite{WINN} followed by a detailed mathematical explanation of our approach. In particular, we focus on explaining how our model generates unseen examples that complement the training data.

\subsection{Background: WINN}
WINN is a recently proposed model that enables generation and discrimination with one single CNN classifier. It incorporates Wasserstein distance into INs to improve the quality of generated samples as well as classification performance. WINN is trained by iteratively sampling self-generated pseudo-negative samples which become increasingly difficult to distinguish from training (positive) samples over the course of training.

\textbf{Discriminative step}
Let us denote training (positive) samples as $S^{+} = \{x_i^+, y_i = +1\}, i = 1, \cdots, N$ and pseudo-negative samples at iteration $t$ as $S^{-}_{t} = \{x_i^-, y_i = -1\}, i = 1, \cdots, N$. The discriminative step trains a CNN classifier $f_{\phi_t}$ parameterized by $\phi_t$ that approximates the Wasserstein distance between positive samples and negative samples at $t$-th iteration. WINN has a gradient penalty term~\cite{wgan-gp}
\begin{equation}
    \mathbb{E}_{\hat{x} \in \hat{S}_t}[\|\nabla_{\hat{x}} f_{\phi_t}(\hat{x})\|_2 - 1]^2,
\end{equation}
where $\hat{S}_t=\{\hat{x_i}\}, i = 1 \cdots, N$ with $\hat{x_i} = \mu_i x_i^+ + (1 - \mu_i) x_i^-$ and $\mu_i$ is the random number drawn from uniform distribution $U[0, 1]$. \\
The objective function of the discriminative step is
\begin{equation}
\begin{split}
    L(\phi_t) = -[\mathbb{E}_{x^+ \in S^{+} }f_{\phi_t}(x^{+}) -  \mathbb{E}_{x^- \in S^{-}_t }f_{\phi_t}(x^{-})] + \\
    \lambda_d \ \mathbb{E}_{\hat{x} \in \hat{S}_t}[\|\nabla_{\hat{x}} f_{\phi_t}(\hat{x})\|_2 - 1]^2,
\end{split}
\end{equation}
where $\lambda_d$ is the weighting factor for the gradient penalty term. This encourages the model to assign a higher score to positive examples than pseudo-negatives while regularizing the disciminative score function $f_{\phi_t}$.

\textbf{Generative step} WINN follows the same generative methods introduced in~\cite{tu2007learning, lazarow2017introspective}. Given a set of positive samples $S^{+}$, WINN iteratively updates self-generated pseudo-negative samples $S^{-}$ to to move them closer to the positive distribution. To directly obtain fair samples from pseudo-negative distribution efficiently, ~\cite{lazarow2017introspective, ICN, WINN} interpret the discriminative model as an estimate of the likelihood ratio
\begin{equation}
    \label{eqn: transform INN}
    \frac{p(x \vert y = +1,\phi_t)}{p(x | y = -1, \phi_t)} \propto
    \frac{p(y = +1 \vert x, \phi_t)}{p(y = -1 \vert x,\phi_t)} = \exp(f_{\phi_t}(x))
\end{equation}
At iteration $t$, the distribution of pseudo-negatives is written as:
\begin{equation}
\label{original update}
    p_{\phi_t}^{-}(x) = \frac{1}{Z_t} \exp(f_{\phi_t}(x)) p_{0}^{-}(x),
\end{equation}
where $p_{\phi_t}^{-}(x)$ represents the negative distribution at $t$-th iteration, $Z_t = \int\exp(f_{\phi_t}(x)) p_{0}^{-}(x)$ is the normalizing factor, and $p_{0}^{-}(x)$ is some initial negative distribution (e.g. a Gaussian). WINN then updates the set of initial pseudo-negative samples to increase $f_{\phi_t}(x)$ via backpropagation. The update of pseudo-negative samples is given by
\begin{equation}
    \nabla x = \frac{\lambda_g}{2} \nabla f_{\phi_t}(x) + \eta,
\end{equation}
where $\lambda_g$ is a time-varying step size and $\eta \sim N(0, \lambda_g)$ is the random Gaussian noise.

\subsection{Resisting Variations via Introspective Learning}

WINN shows promising classification performance by adopting the reclassification-by-synthesis algorithm proposed by ICN~\cite{ICN}. However, both of them fail to capture large data variations between training and testing data since most of the generated pseudo-negatives are very similar to original samples. But in practice, it is very common that testing data contains unseen variations that are not in training data, such as the same objects viewed from different angles and suffered from shape deformations.

To address this issue, we present our approach that utilizes WINN to resist large data variations between training and testing data. Our goal is to produce beneficial unseen variations on the fly and train classifiers with self-generated unseen variations as well as original samples. We assume that we can produce such training samples by applying a transformation function to original training samples. Transformation functions only change the high-level geometric properties of the samples, which ensures the generated samples stay in the same category as original samples. Additionally, we use WINN to generate pseudo-negative samples that approximate transformed positive samples to further enrich the unseen variations. The training steps of ITN are shown in Algorithm 1. We will describe ITN in details in the following paragraphs with consistent notations from the previous section.

\begin{algorithm}[t]
\caption{Outline of ITN Training Algorithm}
\label{euclid}
\begin{algorithmic}[1]
\State \textbf{Input: }Positive sample set $S^{+}$, initial reference distribution $p_{0}^{-}(x)$ and transformation function $\mathcal{T}$
\State \textbf{Output: }Parameters $\theta$, $\phi$ and $\psi$
\State Build $S^-_0$ by sampling $|S^+|$ pseudo-negatives samples from $p_{0}^{-}(x)$
\State initialize parameters $\theta$, $\phi$ and $\psi$, set t = 1
\While {not converge}
    \For {each $x_i^+ \in S^+$ and $x_i^- \in S_t^-$}
    \State Compute transformation parameters $\sigma_i = g_{\psi}(x_i^+)$
    \State Choose $\epsilon_i \sim U(0, 1)$ and compute $\hat{x}_i = \epsilon_i \mathcal{T}(x^{+}_i; \sigma_i) + (\text{1} - \epsilon_i) x^{-}_i$
    \EndFor
    \State Update $\theta, \phi$ with fixed $\psi$ by Eqn.~\ref{eqn: min max}
    \State Update $\psi$ with fixed $\theta, \phi$ by Eqn.~\ref{eqn: min max}
    \State Sample pseudo-negatives samples $Z_t = \{z^t_i, i = 1,...,|S^+|\}$ from $p_{0}^{-}(x)$
    \State Update all samples in $Z_t$ by Eqn.~\ref{eqn: generative_update}
    \State Augment pseudo-negatives sample set $S_t^{-} = S_{t-1}^{-} \cup \{(z^t_i, -1), i=1,...,|S^+|\}$ and t = t + 1
\EndWhile
\end{algorithmic}
\end{algorithm}

\textbf{Enhancing classifiers} \ Denote the transformation function $\mathcal{T}(\cdot; \sigma)$, where $\sigma$ is the function parameter. Similar to the previous notation, let $x^T \in T(S^+; \sigma)$ represent the transformed positive sample. Our objective function consists of two parts. The first part is to make the classifier robust by correctly classifying any given $x^+$, $x^T$, and $x^-$. The multi-class discriminative model is written as
\begin{equation}
    \label{eqn: multi-class}
    q_{t}(y = k \vert x; \theta_t) = \frac{ \text{exp}(f_{\theta^k_t}(x)) }{\sum\limits_{i=1}^n \text{exp}(f_{\theta^{i}_t}(x))},
\end{equation}
where $\theta^{i}_t$ represents the model parameter for class $i$. The first part of the objective function at iteration $t$ is to minimize
\begin{equation}
    L_c(\theta_t) = \ \mathbb{E}_{(x,y) \sim S^ + \cup S_t^- \cup \mathcal{T}(S^+; \sigma_t)}[-\text{log} \ q_t(y \vert x; \theta_t)].
\end{equation}
Classifiers will obtain strong abilities in resisting unseen variations by training with the extra samples while preserving the ability to correctly classify the original samples.

The second part of the objective function is to ensure the generated pseudo-negative samples approximate the transformed positive samples. Therefore, the second part of the objective function is to minimize
\begin{equation}
\begin{split}
    L_w(\phi_t) = -[\mathbb{E}_{x^T \in \mathcal{T}(S^{+}; \sigma_t) }f_{\phi_t}(x^{T}) -  \\
    \mathbb{E}_{x^- \in S^{-}_t }f_{\phi_t}(x^{-})] + \\
    \lambda_d \ \mathbb{E}_{\hat{x} \in \hat{S}_t}[\|\nabla_{\hat{x}} f_{\phi_t}(\hat{x})\|_2 - 1]^2,
\end{split}
\end{equation}
where $\hat{x} = \mu x^T + (1 - \mu) x^-$ with $x^T \in \mathcal{T}(S^+ ; \sigma), x^- \in S^{-}_t, i = 1 \cdots, N$. $\mu$ is the random number drawn from uniform distribution $U[0, 1]$. The total objective function is to minimize
\begin{equation}
    L_w(\phi_t) + L_c(\theta_t). 
\end{equation}

\textbf{Searching transformation parameters} \ However, one problem encountered in the training process is that the transformation function parameter $\sigma_t$ is unknown and there are a huge number of possibilities for selecting $\sigma_t$. Now, the problem becomes how do we learn the $\sigma_t$ in a principled manner and apply it towards building robust classifiers? We solve this issue by formulating a min-max problem:
\begin{equation}
    \label{eqn: min max}
     \min_{\theta, \phi} \ \max_{\sigma} \ L_w(\phi, \sigma) + L_c(\theta, \sigma),
\end{equation}
Here, we rewrite $L_w(\phi_t)$ and $L_c(\theta_t)$ as $L_w(\phi, \sigma)$ and $L_c(\theta, \sigma)$, since $\sigma$ is now an unknown variable. We also subsequently drop the subscript $t$ for notational simplicity. The inner maximization part aims to find the transformation parameter $\sigma$ that achieves the high loss values. The outer minimization is expected to achieve best classification performance as well as minimizing the Wasserstein distance between transformed samples and pseudo-negative samples. However, directly solving Eqn.~\ref{eqn: min max} is very difficult. Thus, we break up this joint optimization into a two step learning process. We first minimize the Eqn.~\ref{eqn: min max} with fixed $\sigma$ to learn $\theta$ and $\phi$. The transformation parameter $\sigma$ is then optimized with previously learned $\theta$ and $\phi$.

More specifically, the parameter $\sigma$ is designed to be a sample dependent parameter. In other words, we want to find a unique transformation for each incoming positive samples. Following this idea, $\sigma$ is learned with function $g_{\psi}(\cdot)$. We have $\sigma = g_{\psi}(x) + \zeta$, where $\zeta$ is random noise follows the standard normal distribution. Notably, following the standard backpropagation procedure, we need to compute the derivative of the transformation function $\mathcal{T}$ in each step. In other words, the transformation function $\mathcal{T}(\cdot; \sigma)$ needs to be differentiable with respect to the parameter $\psi$ to allow gradients flow through the transformation function $\mathcal{T}$ when learning by backpropagation. 

\textbf{Generating pseudo-negative samples} \ ITN generates pseudo-negative samples to further enrich the coverage of unseen variations. We follow the same generation procedure as WINN. Starting from randomly initialized pseudo-negative samples, the update formula is
\begin{equation}
    \label{eqn: generative_update}
    x_{n} = x_{n-1} + \frac{\lambda_g}{2} \nabla f_{\phi_t}(x)) + \eta,
\end{equation}
where $x_{n}$ is the pseudo-negative samples at $n$-th iteration, $\lambda_g$ is the step size, and $\eta \sim N(0, \lambda_g)$ is the random Gaussian noise. An update threshold $T_u$ is introduced to guarantee that the generated pseudo-negative samples are above certain criteria, which prevents bad samples from degrading the classifier performance. We modify the update threshold $T_u$ proposed in~\cite{WINN} and keep track of the $f_{\phi_t}(x)$ in every iteration. In particular, we build a set $D$ by recording $\mathbb{E} [f_{\phi_t}(x)]$, where $x \in S^{+}$ in every iteration. We form a normal distribution $ \mathcal{N}(a, b)$, where $a$ and $b$ represents mean and standard deviation computed from set $D$. The stop threshold is set to be a random number sampled from this normal distribution. The intuition behind this threshold is to make sure the generated negative images are close to the majority of transformed positive images in the feature space.

\section{Experiments}
In this section, we demonstrate the ability of our algorithm to resist large variations between training and testing data through a series of experiments. First, we will present a series of analyses of resisting ability and resisting flexibility of ITN. Following that we demonstrate the strong classification performance of ITN on a series of datasets, including MNIST, affNIST, SVHN, CIFAR-10 and a more challenging dataset, miniImageNet. Finally, we have a discussion section focused on our model effectiveness and some design choices.

\textbf{Experiment Setup} \ Following the setup used in WINN \cite{WINN}, all experiments are conducted with a simple CNN architecture \cite{WINN} unless otherwise specified. We name this simple CNN architecture, B-CNN for notational simplicity. B-CNN contains 4 convolutional layers, each having a $5 \times 5$ filter size with 64 channels and stride 2 in all layers. Each convolutional layer is followed by a batch normalization layer \cite{BN} and a swish activation function \cite{swish}. The last convolutional layer is followed by two consecutive fully connected layers to compute logits and Wasserstein distances. The optimizer used is the Adam optimizer \cite{adam} with parameters $\beta_1 = 0$ and $\beta_2 = 0.9$. 

Our method relies on the transformation function $\mathcal{T}(\cdot)$ to convert original samples to unseen variations. In the following experiments, we demonstrate the ability of ITN to resist large variations with spatial transformers (STs) \cite{jaderberg2015spatial} as our transformation function unless specified. STs includes affine transformations, which are the most common unseen variations in most cases. More importantly, STs are fully differentiable, which allows learning by standard backpropagation.

We compare ITN with other baselines + standard data augmentation unless specified. The standard data augmentation includes the same types of affine transformations as STs, such as rotation, translation, scaling, and shear. The range of data augmentation parameters is set to be same as the range of learned $\sigma$ in Eqn.~\ref{eqn: min max}, which ensures the fair comparison between ITN and data augmentation. In other words, the total number of possible outcome transformations from standard data augmentation and STs are set to be same.

\textbf{Baselines} \ B-CNN mentioned above is our baseline model in this section. Following \cite{ICN} and \cite{WINN}, we compare our method against B-CNN, DCGAN \cite{DCGAN}, WGAN-GP \cite{wgan-gp}, ICN \cite{ICN} and WINN \cite{WINN}. Since DCGAN and WGAN-GP are generative models, to apply them for classification, we adopt the same strategy used in \cite{ICN}. This strategy makes the training phase become a two-step implementation. It first generates negative samples with the original implementation. Then, the generated negative images are used to augment the original training set. Eventually, classifiers are trained with the augmented training set. We denote these two GAN based classification method DCGAN+B-CNN and WGAN-GP+B-CNN. Other methods with B-CNN are denoted as ICN (B-CNN), WINN (B-CNN) and ITN (B-CNN). All results reported in this section are the average of multiple repetitions.

\begin{figure}[h]
\begin{center}
    \includegraphics[scale=0.9]{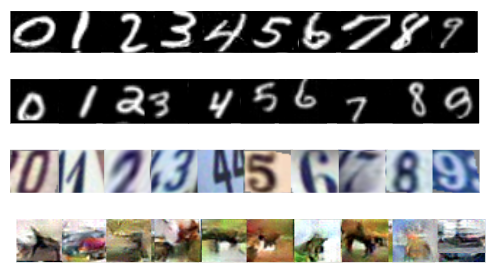}
\end{center}
    \caption{Images generated by ITN. Each row from top to bottom represents the images generated on MNIST, affNIST, SVHN and CIFAR-10.}
    \label{fig: generatedImages}
\end{figure}
\vspace{-10pt}

\subsection{Resisting Large Data Variations}
\textbf{Training-MNIST-Testing-affNIST (TMTA)} We first design a challenging classification task TMTA to verify the ability to resist large data variations of ITN. The training set in this experiment is the MNIST training data while the testing set is the affNIST testing data. MNIST is a benchmark dataset that includes 55000, 5000 and 10000 handwritten digits in the training, validation and testing set, respectively. The affNIST dataset is a variant from the MNIST dataset and it is built by applying various affine transformations to the samples in MNIST dataset. To be consistent with the MNIST dataset and for the following experiments, we reduce the size of training, validation, and testing set of the affNIST dataset to 55000, 5000 and 10000, respectively.
\vspace{-10pt}
\begin{table}[H]
\footnotesize
\begin{center}
\begin{tabular}{c|c}
Method & Error \\
\hline
B-CNN (/w DA) & 40.74\%  \\
DCGAN+B-CNN (/w DA) &  38.51\%  \\
WGAN-GP+B-CNN (/w DA) &  36.29\%  \\
ICN (B-CNN) (/w DA) & 35.79\% \\
WINN (B-CNN) (/w DA) & 33.53\%  \\
\hline
ITN (B-CNN) & \bf{31.67}\%  \\
ITN (B-CNN)(/w DA) & \bf{21.31}\% \\
\end{tabular}
\end{center}
\caption{Testing errors of TMTA task.}
\label{tab: TMTA}
\end{table}

Even though the training and testing samples are from ``different datasets'', we believe it is reasonable to consider the training and testing samples are from the same underlying distribution. From this perspective, the difficulty of this task is how to overcome such huge data variations between training and testing set since the testing set includes much more variations. As shown in Table \ref{tab: TMTA}, ITN outperforms other methods with standard data augmentation. 

\textbf{Limited training data} Another way to evaluate the ability of resisting variations is to reduce the number of training samples. Intuitively, data variations between training and testing sets become larger when the number of testing data remains the same while the number of samples in the training set shrinks. Thus, we implicitly increase the variations between the training and testing data by reducing the number of samples in the training data. The purpose of this experiment is to demonstrate the potential of ITN to resist unseen variations from a different perspective.
\vspace{-10pt}
\begin{table}[H]
\footnotesize
\begin{center}
\begin{tabular}{l|c|c|c|c}
Method & 0.1\% & 1\% & 10\% &25\% \\
\hline
B-CNN (w/ DA) & 18.07\% & 4.48\% & 1.24\% & 0.83 \% \\

DCGAN+B-CNN (w/ DA) & 16.17\% & 4.13\% & 1.21\% &0.81 \% \\

WGAN-GP+B-CNN (w/ DA) & 15.35\% & 3.98\% & 1.18\% &0.79 \% \\

ICN (B-CNN) (w/ DA) & 15.12\% & 3.74\% & 1.09\% & 0.80 \% \\

WINN (B-CNN) (w/ DA) & 14.64\% & 3.66\% & 1.00\% & 0.77 \% \\
\hline
ITN (B-CNN) & \bf{12.85}\% & \bf{3.18}\% & \bf{0.93}\% & \bf{0.73}\%  \\

ITN (B-CNN) (w/ DA) & \bf{11.57}\% & \bf{2.78}\% & \bf{0.89}\% & \bf{0.65}\%  \\
\end{tabular}
\end{center}
\caption{Testing errors of the classification results with limited training data, where 0.1\% means the training data is randomly selected 0.1\% of the MNIST training data while the testing data is the entire MNIST testing data.}
\label{tab: few-shot}
\end{table}

We design a new experiment where the training set is the MNIST dataset with only 0.1\%, 1\%, 10\% and 25\% of the whole training set while the testing set is the \textit{entire} MNIST testing set. The reduced training set is built by randomly sampling data from the MNIST training data while keeping the number of data per class identical. As shown in Table~\ref{tab: few-shot}, our method has better results on all tasks, which are consistent with our previous results. The constantly superior performance of ITN over data augmentation indicates its effectiveness.

\textbf{Beyond spatial transformer} Even though we utilize STs to demonstrate our ability to resist data variations, our method can generalize to other types of transformations. Our algorithm can take other types of differentiable transformation functions and similarly strengthen discriminators. Moreover, our algorithm can utilize multiple types of transformation functions at the same time and provide an even stronger ability to resist mixed variations simultaneously. To verify this, we introduce another recently proposed work, Deep Diffeomorphic Transformer (DDT) Networks \cite{ddt, cpab}. DDTs are similar to STs in a way that both of them are optimized through standard backpropagation.

We replace ST modules with DDT modules and check whether our algorithm can resist such type of variation. Then, we include both STs and DDTs in our model and verify the performance again. Let MNIST dataset be the training set of the experiment while the testing set is the MNIST dataset with different types of transformation applied. We introduce two types of testing sets in this section. The first one is the normal testing set with random DDT transformations only. The second one is similar to the first one but includes both random DDT and affine transformations. The DDT transformation parameters are drawn from $N(0, 0.7 \times \mathcal{I}_d)$ as suggested in \cite{ddt}, where $\mathcal{I}_d$ represents the $d$ dimensional identity matrix. Then the transformed images are randomly placed in a $42 \times 42$ images. Lastly, we replicate the same experiment on the CIFAR-10 dataset.
\vspace{-10pt}
\begin{table}[h]
\scriptsize
\begin{center}
\begin{tabular}{l|c|c|c|l}
{} & \multicolumn{2}{c|}{MNIST} & \multicolumn{2}{c}{CIFAR-10} \\
\hline
{} & DDT & DDT + ST & DDT & {DDT + ST} \\
\hline
B-CNN & 17.75\% & 55.11\% & 76.14 \%& 78.01 \%\\
WGAN-GP+B-CNN & 17.53\% & 53.24\%& 75.93\% & 77.02 \%\\
WINN (B-CNN) & 17.20\% & 52.43\% & 75.43 \%& 76.92 \%\\
\hline
ITN-V1 (B-CNN) & 12.85 \% & 40.60\% & 53.62\% &63.56 \% \\
ITN-V2 (B-CNN) & \bf{9.41\%} & \bf{34.37\%} & \bf{45.26\%} & \bf{56.95 \%}\\
\end{tabular}
\end{center}
\caption{Testing errors of classification results under different testing data transformations. ITN-V1 represents ITN with DDT transformation function and ITN-V2 represents ITN with DDT and ST transformation functions together. }
\label{tab: beyond}
\end{table}

\textit{Agnostic to different transformation functions} We observe from Table~\ref{tab: beyond} that ITN-V1 (B-CNN) improves the discriminator performance by 4.35\% from WINN (B-CNN) on MNIST with random DDT transformations and by 21.81\% on CIFAR-10 dataset. In other words, ITN successfully resists DDT type of variations by integrating with DDT transformation function. Together with results from Table \ref{tab: MNIST & affNIST}, we see that ITN can work with different types of transformation function and resists the corresponding type of variations. 

\textit{Integrating multiple transformation functions} Another important observation from Table \ref{tab: beyond} is that ITN-V2 (B-CNN) can utilize multiple transformations at the same time to resist a mixture of corresponding variations. Comparing against ITN-V1 (B-CNN), ITN-V2 (B-CNN) reduces the testing errors by 6.23\% on MNIST dataset with random DDT + ST type of variations. Additionally, it reduces the testing errors by 6.61\% on CIFAR-10 dataset.

More importantly, the performance of ITN does not degrade when the model has transformation functions that do not match the type of variations in the testing data, e.g. ITN-V2 (B-CNN) on testing data with DDT only. From this observation, we conclude that applying extra transformations functions in the ITN will not degrade the performance even though the testing data does not have such transformations. The reason for this observation is most likely that ITN generates different transformations in every iteration, which helps it avoid over-reliance on a particular transformation. After correctly classifying original samples and transformed samples, ITN can model more complicated data distributions than other methods.

\subsection{Classification}
\textbf{MNIST and affNIST} Now we want to back to the most common task, classification to ensure ITN not only work well on large data variations but also boost performance on well-known benchmark datatsets. We first compare the performance of ITN against other baselines on MNIST and affNIST datasets and then gradually switch to harder datasets. As shown in Table~\ref{tab: MNIST & affNIST}, ITN outperforms other baselines. It is easy to observe that the performance improvements on MNIST and affNIST datasets are marginal compared to experiments in Section 4.1. The most likely explanation for this observation is that the training samples in MNIST and affNIST represent the data distribution very well. In this case, providing more samples will not significantly boost the performance of the classifier.
\vspace{-10pt}
\begin{table}[h]
\footnotesize
\begin{center}
\begin{tabular}{l|c|c}
Method & MNIST & AffNIST \\
\hline
B-CNN (/w DA) & 0.57\% &1.65\% \\
DCGAN+B-CNN (/w DA) & 0.57\% & 1.63\% \\
WGAN-GP+B-CNN (/w DA) & 0.56\% & 1.56\% \\
ICN (B-CNN) (/w DA)) & 0.56\% &1.54\% \\
WINN (B-CNN) (/w DA) & 0.52\% &1.48\% \\
\hline
ITN (B-CNN) & \bf{0.47\%} &\bf{1.42\%} \\
ITN (B-CNN) (/w DA) & \bf{0.42\%} &\bf{1.09\%} \\
\end{tabular}
\end{center}
\caption{Testing errors on MNIST and affNIST, where /w DA represents the method is trained with standard data augmentation.}
\label{tab: MNIST & affNIST}
\vspace{-5pt}
\end{table}

\textbf{SVHN and CIFAR-10}
Next, we evaluate the performance of ITN on SVHN and CIFAR-10 datasets. SVHN~\cite{svhn} is a dataset that contains house number images from Google Street View. There are 73257 digits for training, 26032 digits for testing in SVHN dataset. The CIFAR-10 dataset~\cite{cifar} consists of 60000 color images of size 32 $\times$ 32. This set of 60000 images is split into two sets, 50000 images for training and 10000 images for testing. In this section, we also use ResNet-32~\cite{ResNet} as a baseline backbone to validate the performance of our framework with deeper network architectures, following the setting in~\cite{WINN}. ITN outperforms other methods on SVHN and CIFAR-10 datasets as shown in Table \ref{tab: svhn and cifar}. Some samples generated by ITN are shown in Figure \ref{fig: generatedImages}.
\vspace{-10pt}
\begin{table}[H]
\footnotesize
\begin{center}
\begin{tabular}{c|c|c}
Method & SVHN & CIFAR-10 \\
\hline
B-CNN (w/DA) & 7.01\% & 24.35\% \\
ResNet-32 (w/DA) & 4.03\% & 7.51\% \\
DCGAN + ResNet-32 (w/DA) & 3.87\% & 7.17\% \\
WGAN-GP + ResNet-32 (w/DA) & 3.81\% & 7.05\% \\
ICN (ResNet-32) (w/DA) & 3.76\% & 6.70\% \\
WINN (ResNet-32) (w/DA) & 3.68\% & 6.43\% \\
\hline
ITN (ResNet-32) & \bf{3.47\%} & \bf{6.08\%} \\
ITN (ResNet-32) (w/DA) & \bf{3.32\%} & \bf{5.82\%} \\
\end{tabular}
\end{center}
\caption{Testing errors on SVHN and CIFAR-10.}
\label{tab: svhn and cifar}
\end{table}
\textbf{A more challenging dataset -- miniImageNet} We further verify the scalability of ITN by evaluating our proposed method on a new dataset named miniImageNet~\cite{matching, ren2018meta}. MiniImageNet dataset is a modified version of the ILSVRC-12 dataset~\cite{russakovsky2015imagenet}, in which 600 images for each of 100 classes were randomly chosen to be part of the dataset. All images in this dataset are of size $84 \times 84$ pixels. Compared to previously tested datasets in this section, miniImageNet is significantly harder choice both in terms of generation and classification. Each sample contains complicated natural scenes that pose a hard challenge in generating pseudo-negative samples. Additionally, the number of classes are also larger than previously seen datasets. 

The results are shown in Table \ref{tab: miniImageNet} and ITN shows consistent better performance than all other comparisons. Note that the performance improvement of ITN on miniImagenet is slightly lower than expected. One possible reason is that the generative ability of WINN is still limited on this challenging dataset. Additionally, the generation speed of pseudo-negatives are about 6 times slower compared to the CIFAR-10 dataset due to the complexity of samples. 
\vspace{-10pt}
\begin{table}[h]
\footnotesize
\begin{center}
\begin{tabular}{c|c}
Method & Error \\
\hline
ResNet-32 (w/DA) & 35.25\% \\
DCGAN + ResNet-32 (w/DA) & 33.06\% \\
WGAN-GP + ResNet-32(w/DA) & 33.42\% \\
ICN (ResNet-32) (w/DA) & 32.87\% \\
WINN (ResNet-32) (w/DA) & 32.18\% \\
\hline
ITN (ResNet-32) & \bf{31.56\%} \\
ITN (ResNet-32) (w/DA) & \bf{29.65\%} \\
\end{tabular}
\end{center}
\caption{Testing errors on the miniImageNet dataset.}
\label{tab: miniImageNet}
\end{table}

\vspace{-10pt}
\subsection{Discussion}
\textbf{Against data augmentation} \ Comparing ITN against data augmentation, they have their unique advantages and disadvantages. ITN outperforms data augmentation given all results reported in previous sections. It is also a well-formulated model compared to an exhaustively searching method like data augmentation. However, data augmentation is faster than ITN, especially when the dataset becomes large and complicated. Note that ITN works well with data augmentation as ITN (w/DA) produces the best performance. This observation adds more practical value to ITN because they can be jointly applied without contradictions. 

\begin{figure}[h]
\begin{center}
    \includegraphics[scale=0.42]{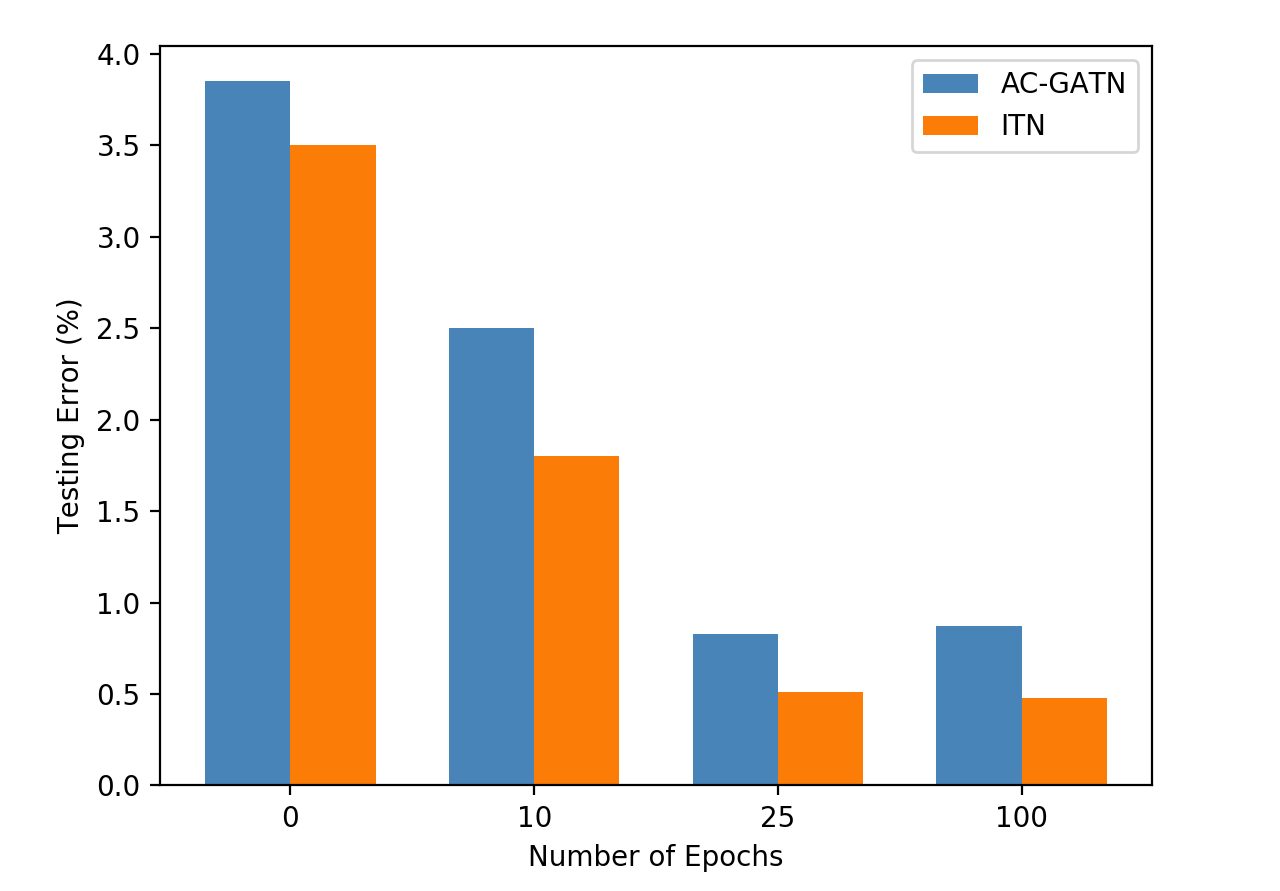}
\end{center}
    \caption{Testing errors of AC-GATN (B-CNN) and ITN (B-CNN) on the MNIST dataset.}
    \label{fig: compare performance}
\end{figure}

\begin{figure}[h]
\begin{center}
    \includegraphics[scale=0.38]{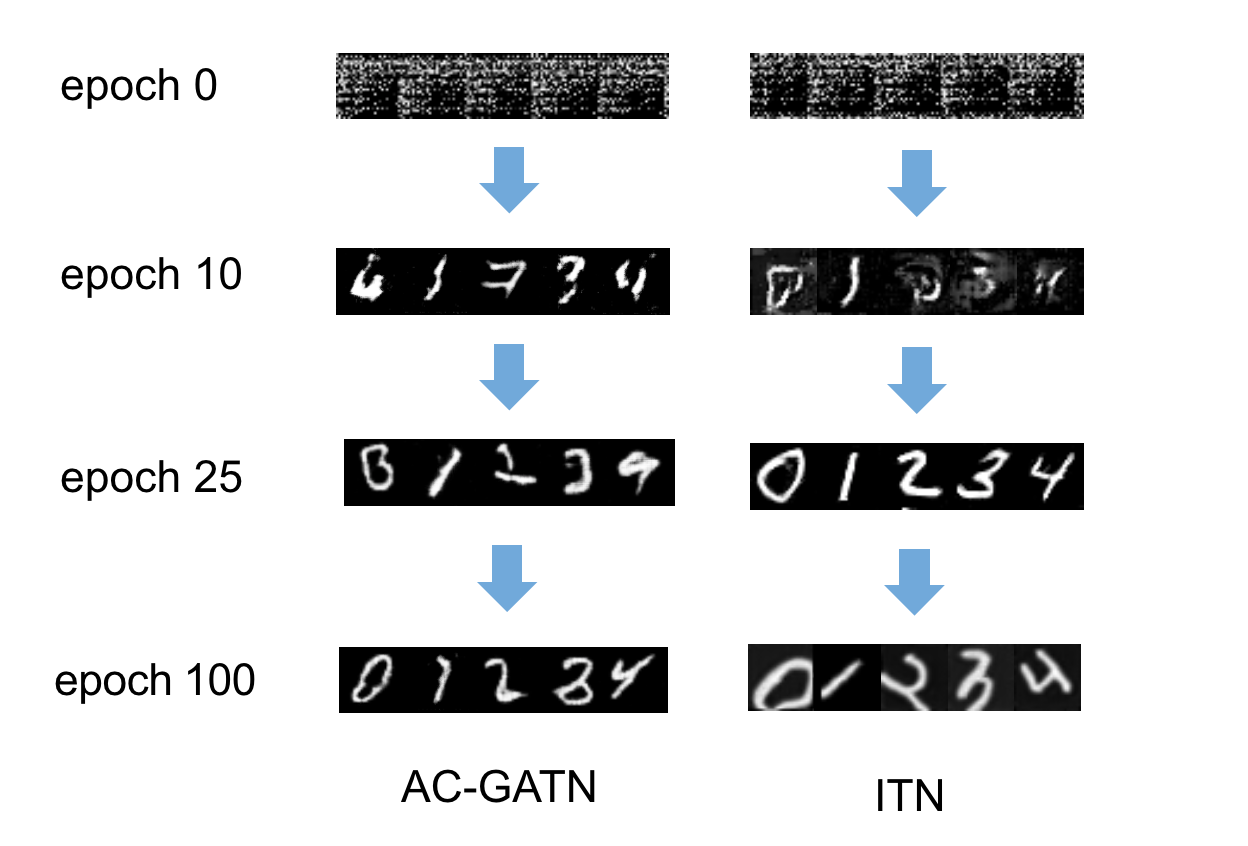}
\end{center}
    \caption{Samples generated by AC-GATN (B-CNN) and ITN (B-CNN) on MNIST.}
    \label{fig: compare samples}
\end{figure}

\textbf{Choice w.r.t generative models} \ It is essential to validate our choice of generative models. We implement our framework by using AC-GAN as the generative model and name it \textit{AC-GATN}. AC-GAN can generate class-dependent samples, which is required in our framework. The loss function of AC-GAN is replaced with the loss function from WGAN-GP to directly compare it with ITN. All experimental settings are the same for a fair comparison. The results shown in Figure \ref{fig: compare performance} illustrate that under our framework, using INs as the generative model achieve better performance than using GANs. By visualizing the generated samples from AC-GATN and ITN (shown in Figure \ref{fig: compare samples}), both AC-GATN and ITN generate clear and sharp images. However, samples generated by AC-GATN have lower quality on average in terms of human standards as some of them are misleading and inaccurate, i.e. the number 3 is close to number 6 in epoch 100. These lower quality samples will mislead the classifier and lead to a performance decrease. Consequently, we choose INs rather than GANs in our approach. We will provide more comparisons of AC-GATN and ITN on other datasets in the supplementary material.

\vspace{-15pt}

\begin{figure}[h]
\begin{center}
    \includegraphics[scale=0.5]{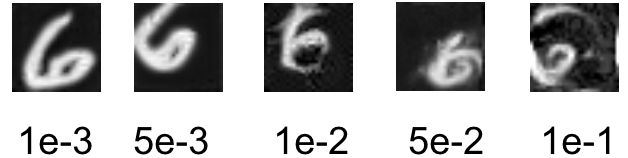}
\end{center}
    \caption{Samples generated by ITN with different thresholds $T_u$. The number below each sample represents the threshold. }
    \label{fig: threshold}
\end{figure}

\textbf{Effects of the update threshold $\mathbf{T_u}$} \ The update threshold $T_u$ introduced in INs quantitatively controls the quality of samples in the generation process. In Table~\ref{table:threshold}, we present the results of ITN on the MNIST dataset with different thresholds to explore the relationship between the samples quality and thresholds. Not surprisingly, we observe that the performance of ITN drops when increasing the threshold. By visualizing the samples generated by different thresholds, it is clear that the performance drops due to the decrease in the quality of generated samples (Fig.~\ref{fig: threshold}). Although our performance drops with the increase of the threshold, in a certain range ($<5e-3$), our result is still better than others shown in Table~\ref{tab: MNIST & affNIST}, which shows that our approach tolerates samples of low qualities in some extend.

\begin{table}[H]
\small
\begin{center}
\begin{tabular}{c|c|c|c|c|c}
\hline
$T_u$ & 1e-3 (default) & 5e-3 & 1e-2 & 5e-2 & 1e-1 \\
\hline
ITN error & 0.47\% & 0.51\% & 0.67\% & 0.78\% & 0.92\% \\
\hline
\end{tabular}
\end{center}
\caption{Testing errors of ITN (B-CNN) with various thresholds on MNIST.}\label{table:threshold}
\end{table}
\vspace{-10pt}

\textbf{Effects of pseudo-negative samples} \ It is worth exploring how pseudo-negative samples affect the performance of the classifier. Pseudo-negative samples are considered as samples generated from the transformed positive distribution. Therefore, having a pool of pseudo-negative samples that approximate transformed positive samples will cover a wider range of unseen variations. A simple evaluation is to check the performance of ITN without the generation module. In this case, the training objective will be correctly classifying original and transformed positive samples. We evaluate the performance of this baseline on MNIST dataset, affNIST dataset, and TMTA task. We observe from Table~\ref{tab: adversarial} that ITN-NG has the same performance as ITN on simple dataset while it has lower performance than ITN when the task becomes harder.
\begin{table}[h]
\footnotesize
\begin{center}
\begin{tabular}{c|c|c|c}
{} & MNIST & affNIST & TMTA \\
\hline
ITN-NG & 0.49\% & 1.45\% & 33.02\% \\
\hline
ITN & \textbf{0.47}\% & \textbf{1.42}\% & \textbf{31.67}\% \\
\hline
\end{tabular}
\end{center}
\caption{ Testing errors of ITN and ITN-NG on MNIST, affNIST, and TMTA task, where ITN-NG is the version of ITN without generating pseudo-negative samples.}\label{tab: adversarial}
\end{table}
\vspace{-10pt}

\section{Conclusion}
\vspace{-2mm}
We proposed a principled and efficient approach that endows the classifiers with the ability to resist larger variations between training and testing data. Our method, ITN, strengthens the classifiers by generating unseen variations with various learned transformations. Experimental results show consistent performance improvements not only on the classification tasks but also on the other challenging classification tasks, such as TMTA. Moreover, ITN demonstrates its advantages in both effectiveness and efficiency over data augmentation. Our future work includes applying our approach to large scale datasets and extending it to generate samples with more types of variations.

\noindent\textbf{Acknowledgement} This work was supported by NSFC No. 61672336 and Office of Naval Research N00014-15-1-2356.

{\small
\bibliographystyle{ieee}
\bibliography{egbib}

\begin{thebibliography}{10}\itemsep=-1pt

\bibitem{wgan}
M.~Arjovsky, S.~Chintala, and L.~Bottou.
\newblock Wasserstein gan.
\newblock {\em arXiv preprint arXiv:1701.07875}, 2017.

\bibitem{denton2015deep}
E.~L. Denton, S.~Chintala, R.~Fergus, et~al.
\newblock Deep generative image models using a￼ laplacian pyramid of
  adversarial networks.
\newblock In {\em Advances in neural information processing systems}, pages
  1486--1494, 2015.

\bibitem{ddt}
N.~S. Detlefsen, O.~Freifeld, and S.~Hauberg.
\newblock Deep diffeomorphic transformer networks.
\newblock {\em Conference on Computer Vision and Pattern Recognition (CVPR)},
  2018.

\bibitem{cpab}
O.~Freifeld, S.~Hauberg, K.~Batmanghelich, and J.~W. Fisher.
\newblock Transformations based on continuous piecewise-affine velocity fields.
\newblock {\em IEEE Transactions on Pattern Analysis and Machine Intelligence},
  2017.

\bibitem{friedman2001elements}
J.~Friedman, T.~Hastie, and R.~Tibshirani.
\newblock {\em The elements of statistical learning}, volume~1.
\newblock Springer series in statistics New York, NY, USA:, 2001.

\bibitem{goodfellow2014generative}
I.~Goodfellow, J.~Pouget-Abadie, M.~Mirza, B.~Xu, D.~Warde-Farley, S.~Ozair,
  A.~Courville, and Y.~Bengio.
\newblock Generative adversarial nets.
\newblock In {\em Advances in neural information processing systems}, pages
  2672--2680, 2014.

\bibitem{wgan-gp}
I.~Gulrajani, F.~Ahmed, M.~Arjovsky, V.~Dumoulin, and A.~C. Courville.
\newblock Improved training of wasserstein gans.
\newblock In {\em Advances in Neural Information Processing Systems}, pages
  5769--5779, 2017.

\bibitem{ResNet}
K.~He, X.~Zhang, S.~Ren, and J.~Sun.
\newblock Deep residual learning for image recognition.
\newblock In {\em Proceedings of the IEEE conference on computer vision and
  pattern recognition}, pages 770--778, 2016.

\bibitem{huang2017densely}
G.~Huang, Z.~Liu, K.~Q. Weinberger, and L.~van~der Maaten.
\newblock Densely connected convolutional networks.
\newblock In {\em Proceedings of the IEEE conference on computer vision and
  pattern recognition}, volume~1, page~3, 2017.

\bibitem{BN}
S.~Ioffe and C.~Szegedy.
\newblock Batch normalization: Accelerating deep network training by reducing
  internal covariate shift.
\newblock {\em arXiv preprint arXiv:1502.03167}, 2015.

\bibitem{isola2017image}
P.~Isola, J.-Y. Zhu, T.~Zhou, and A.~A. Efros.
\newblock Image-to-image translation with conditional adversarial networks.
\newblock {\em arXiv preprint}, 2017.

\bibitem{jaderberg2015spatial}
M.~Jaderberg, K.~Simonyan, A.~Zisserman, et~al.
\newblock Spatial transformer networks.
\newblock In {\em Advances in neural information processing systems}, pages
  2017--2025, 2015.

\bibitem{jebara2012machine}
T.~Jebara.
\newblock {\em Machine learning: discriminative and generative}, volume 755.
\newblock Springer Science \& Business Media, 2012.

\bibitem{ICN}
L.~Jin, J.~Lazarow, and Z.~Tu.
\newblock Introspective classification with convolutional nets.
\newblock In {\em Advances in Neural Information Processing Systems}, pages
  823--833, 2017.

\bibitem{adam}
D.~P. Kingma and J.~Ba.
\newblock Adam: A method for stochastic optimization.
\newblock {\em arXiv preprint arXiv:1412.6980}, 2014.

\bibitem{cifar}
A.~Krizhevsky and G.~Hinton.
\newblock Learning multiple layers of features from tiny images.
\newblock 2009.

\bibitem{alexNet}
A.~Krizhevsky, I.~Sutskever, and G.~E. Hinton.
\newblock Imagenet classification with deep convolutional neural networks.
\newblock In {\em Advances in neural information processing systems}, pages
  1097--1105, 2012.

\bibitem{lazarow2017introspective}
J.~Lazarow, L.~Jin, and Z.~Tu.
\newblock Introspective neural networks for generative modeling.
\newblock In {\em Proceedings of the IEEE Conference on Computer Vision and
  Pattern Recognition}, pages 2774--2783, 2017.

\bibitem{CNN}
Y.~LeCun, B.~Boser, J.~S. Denker, D.~Henderson, R.~E. Howard, W.~Hubbard, and
  L.~D. Jackel.
\newblock Backpropagation applied to handwritten zip code recognition.
\newblock {\em Neural computation}, 1(4):541--551, 1989.

\bibitem{WINN}
K.~Lee, W.~Xu, F.~Fan, and Z.~Tu.
\newblock Wasserstein introspective neural networks.
\newblock In {\em Proceedings of the IEEE Conference on Computer Vision and
  Pattern Recognition}, 2018.

\bibitem{liang2008asymptotic}
P.~Liang and M.~I. Jordan.
\newblock An asymptotic analysis of generative, discriminative, and
  pseudolikelihood estimators.
\newblock In {\em Proceedings of the 25th international conference on Machine
  learning}, pages 584--591. ACM, 2008.

\bibitem{cgan}
M.~Mirza and S.~Osindero.
\newblock Conditional generative adversarial nets.
\newblock {\em arXiv preprint arXiv:1411.1784}, 2014.

\bibitem{svhn}
Y.~Netzer, T.~Wang, A.~Coates, A.~Bissacco, B.~Wu, and A.~Y. Ng.
\newblock Reading digits in natural images with unsupervised feature learning.
\newblock In {\em NIPS workshop on deep learning and unsupervised feature
  learning}, volume 2011, page~5, 2011.

\bibitem{acgan}
A.~Odena, C.~Olah, and J.~Shlens.
\newblock Conditional image synthesis with auxiliary classifier gans.
\newblock In {\em Proceedings of the 34th International Conference on Machine
  Learning-Volume 70}, pages 2642--2651. JMLR. org, 2017.

\bibitem{DCGAN}
A.~Radford, L.~Metz, and S.~Chintala.
\newblock Unsupervised representation learning with deep convolutional
  generative adversarial networks.
\newblock {\em arXiv preprint arXiv:1511.06434}, 2015.

\bibitem{swish}
P.~Ramachandran, B.~Zoph, and Q.~V. Le.
\newblock Searching for activation functions.
\newblock 2018.

\bibitem{ren2018meta}
M.~Ren, E.~Triantafillou, S.~Ravi, J.~Snell, K.~Swersky, J.~B. Tenenbaum,
  H.~Larochelle, and R.~S. Zemel.
\newblock Meta-learning for semi-supervised few-shot classification.
\newblock {\em arXiv preprint arXiv:1803.00676}, 2018.

\bibitem{russakovsky2015imagenet}
O.~Russakovsky, J.~Deng, H.~Su, J.~Krause, S.~Satheesh, S.~Ma, Z.~Huang,
  A.~Karpathy, A.~Khosla, M.~Bernstein, et~al.
\newblock Imagenet large scale visual recognition challenge.
\newblock {\em International Journal of Computer Vision}, 115(3):211--252,
  2015.

\bibitem{improved-gan}
T.~Salimans, I.~Goodfellow, W.~Zaremba, V.~Cheung, A.~Radford, and X.~Chen.
\newblock Improved techniques for training gans.
\newblock In {\em Advances in Neural Information Processing Systems}, pages
  2234--2242, 2016.

\bibitem{VGG}
K.~Simonyan and A.~Zisserman.
\newblock Very deep convolutional networks for large-scale image recognition.
\newblock {\em arXiv preprint arXiv:1409.1556}, 2014.

\bibitem{googleNet}
C.~Szegedy, W.~Liu, Y.~Jia, P.~Sermanet, S.~Reed, D.~Anguelov, D.~Erhan,
  V.~Vanhoucke, A.~Rabinovich, et~al.
\newblock Going deeper with convolutions.
\newblock CVPR, 2015.

\bibitem{tu2007learning}
Z.~Tu.
\newblock Learning generative models via discriminative approaches.
\newblock In {\em Computer Vision and Pattern Recognition, 2007. CVPR'07. IEEE
  Conference on}, pages 1--8. IEEE, 2007.

\bibitem{tu2008brain}
Z.~Tu, K.~L. Narr, P.~Doll{\'a}r, I.~Dinov, P.~M. Thompson, and A.~W. Toga.
\newblock Brain anatomical structure segmentation by hybrid
  discriminative/generative models.
\newblock {\em IEEE transactions on medical imaging}, 27(4):495--508, 2008.

\bibitem{matching}
O.~Vinyals, C.~Blundell, T.~Lillicrap, D.~Wierstra, et~al.
\newblock Matching networks for one shot learning.
\newblock In {\em Advances in Neural Information Processing Systems}, pages
  3630--3638, 2016.

\bibitem{welling2003self}
M.~Welling, R.~S. Zemel, and G.~E. Hinton.
\newblock Self supervised boosting.
\newblock In {\em Advances in neural information processing systems}, pages
  681--688, 2003.

\bibitem{zhu2017unpaired}
J.-Y. Zhu, T.~Park, P.~Isola, and A.~A. Efros.
\newblock Unpaired image-to-image translation using cycle-consistent
  adversarial networks.
\newblock {\em arXiv preprint}, 2017.

\end{thebibliography}
}

\end{document}